\documentclass[sigconf]{acmart}

\usepackage{xspace}
\usepackage{subcaption}
\usepackage{amsmath, bbm}
\usepackage{wrapfig}
\usepackage{captdef}
\usepackage{arydshln}
\usepackage{comment, makecell}
\usepackage{multirow}
\usepackage{graphicx}
\usepackage{enumitem}
\usepackage{algorithm}
\usepackage{algorithmic}
\usepackage{pifont}

\newlength\savewidth

\usepackage{amsmath, amsthm}

\usepackage{booktabs}
\usepackage{array,multirow}
\usepackage{cleveref}
\usepackage{tabularx}
\usepackage{boldline}
\crefformat{section}{\S#2#1#3}
\crefformat{subsection}{\S#2#1#3}
\crefformat{subsubsection}{\S#2#1#3}

\newlist{researchquestions}{enumerate}{1}
\setlist[researchquestions]{label*=\textbf{RQ\arabic*}}

\AtBeginDocument{%
  \providecommand\BibTeX{{%
    \normalfont B\kern-0.5em{\scshape i\kern-0.25em b}\kern-0.8em\TeX}}}

\setcopyright{acmlicensed}
\copyrightyear{2018}
\acmYear{2018}
\acmDOI{XXXXXXX.XXXXXX}

\acmConference[Woodstock ’18]{Proceedings of ACM Web
Conference.}{June 03--05,
  2018}{Woodstock, NY}
\acmISBN{978-1-4503-XXXX-X/18/06}

\begin{document}

\title{Improving Matrix Completion by Exploiting Rating Ordinality in Graph Neural Networks}


\author{Jaehyun Lee}
\authornote{Both authors contributed equally to this research.}
\affiliation{%
  \institution{POSTECH}
  \country{South Korea}
}
\email{jminy8@postech.ac.kr}

\author{SeongKu Kang}
\authornotemark[1]
\affiliation{%
  \institution{UIUC}
  \country{USA}
}
\email{seongku@illinois.edu}

\author{Hwanjo Yu}
\authornote{Hwanjo Yu is the corresponding author}
\affiliation{%
  \institution{POSTECH}
  \country{South Korea}
}
\email{hwanjoyu@postech.ac.kr}

\begin{abstract}
Matrix completion is an important area of research in recommender systems.
Recent methods view a rating matrix as a user-item bipartite graph with labeled edges denoting observed ratings and predict the edges between the user and item nodes by using the graph neural network (GNN).
Despite their effectiveness, they treat each rating type as an independent relation type and thus cannot sufficiently consider the ordinal nature of the ratings.
In this paper, we explore a new approach to exploit rating ordinality for GNN, which has not been studied well in the literature.
We introduce a new method, called \proposed,  to leverage \underline{R}ating \underline{O}rdinality in \underline{G}NN-based \underline{M}atrix \underline{C}ompletion.
It uses \textit{cumulative preference propagation} to directly incorporate rating ordinality in GNN's message passing, allowing for users' stronger preferences to be more emphasized based on inherent orders of rating types.
This process is complemented by \textit{interest regularization} which facilitates preference learning using the underlying interest information.
Our extensive experiments show that \proposed consistently outperforms the existing strategies of using rating types for GNN.
We expect that our attempt to explore the feasibility of utilizing rating ordinality for GNN may stimulate further research~in~this~direction.

\end{abstract}


\newcommand{\proposed}{ROGMC\xspace}
\newcommand{\impgraph}{$\mathcal{G}_{I}$\xspace}
\newcommand{\xmark}{\ding{55}}%


\maketitle

\section{Introduction}
Matrix completion is an important problem in recommender systems \cite{MF, chen2018matrix}.
Given a partially observed user-item matrix whose entries represent ratings from users on items, it aims to predict the missing entries in the matrix based on the observed ones. 
Recent studies \cite{inductiveMC_CIKM21, inductiveMC_ICLR20, GCMC, CPA-LGC} have achieved remarkable performance by using the graph neural networks (GNN).
They view the rating matrix as a user-item bipartite graph where observed ratings are represented by labeled edges, then they apply GNN to enrich each node's representation by propagating the information of~multi-hop~neighbors.

An important question is how to reflect rating types of edges for GNN's message passing.
Most GNN-based methods \cite{GCMC, relation_GNN, zhang2019star, inductiveMC_ICLR20, inductiveMC_CIKM21, edge_GNN} have treated each rating as an \textit{independent relation type} and differently propagate the information of neighboring nodes according to their rating types.
The most popular approach is to use relation-wise transformation, which has shown high effectiveness in handling diverse relations of knowledge graph \cite{relation_GNN}. 
Specifically, \cite{GCMC, zhang2019star, inductiveMC_ICLR20} apply the rating-wise transformation that uses different parameter matrices for each rating type in the message passing layer.
On the one hand, \cite{edge_GNN} employs rating-wise edge embedding to distinguish the message passing via edges with different rating types.
Lastly, \cite{inductiveMC_CIKM21} uses rating-wise propagation to separately propagate representations for each rating type.
Commonly, they treat rating types as a kind of categorical label in the sense that they are treated equally and independently~of~each~other.

However, rating types are not inherently independent relations, as they have an \textit{ordinal nature}.
That is, there exist inherent orders among different rating types, i.e., higher ratings generally reflect a user's stronger preferences compared to lower ratings.
Such ordinality among the relation types is an important distinction from other application graphs, such as knowledge graphs, where relations (e.g., citizen\_of, educated\_at) do not possess inherent orders.
In this regard, we argue that the independent modeling of the existing GNN-based methods insufficiently reflects the unique nature of ratings, which may result in suboptimal performance.
Indeed, we observe that the independent modeling often fails to achieve improvements over a simple GNN that completely disregards rating types, particularly when the training data~is~limited~(reported~in~\cref{subsec:exp_result}).

In this paper, we explore a way to exploit rating ordinality for GNN, which has not been studied well in the literature.
In machine learning, a well-known strategy to leverage ordinality involves converting an ordinal attribute into multiple binary attributes based on order relations \cite{frank2001simple}. 
Specifically, an ordinal attribute $R$ with ordered values $(R_{1}, R_{2}, \cdots, R_{k})$ is converted into $k$ binary attributes.
The $i$-th binary attribute is defined as $I[R \geq R_{i}]$, where $I$ denotes the indicator function. 
As a concrete example, consider an ordinal attribute $R$ with three values: $(Cool, Mild, Hot)$.
This can be converted into three binary attributes, i.e., $I[R \geq Cool]$, $I[R \geq Mild]$, and $I[R \geq Hot]$.
The converted binary attributes are used for the subsequent training process.
This strategy allows for naturally considering the inherent orders among the ordinal values, thus better reflecting the ordinality compared to treating them independently~\cite{frank2001simple}.

\begin{figure}[t!]
\centering
    \includegraphics[width=1.\linewidth]{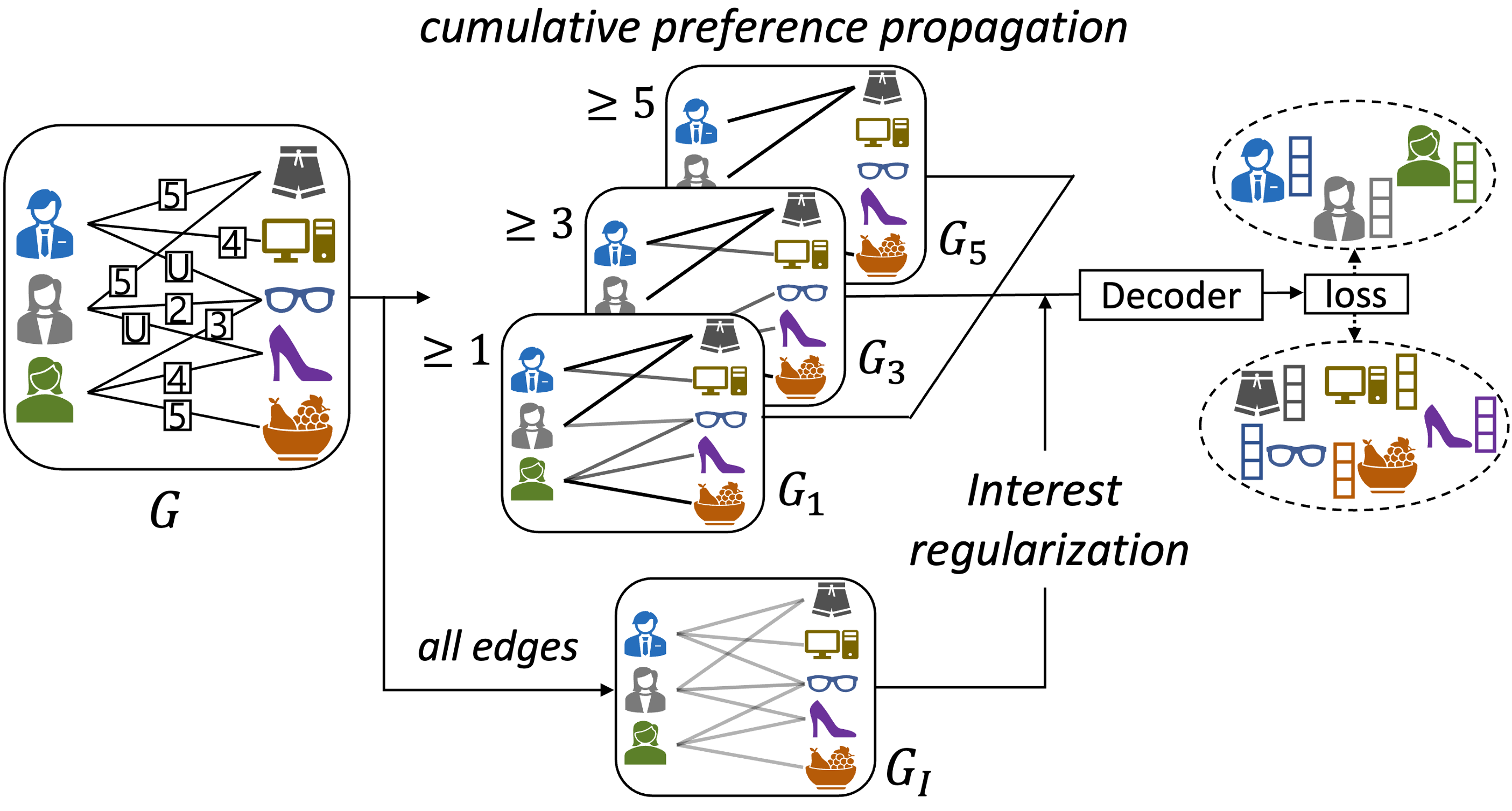}
    \caption{Method overview. `U' denotes observed interaction with no rating from users. This example uses $\mathcal{T}=\{1,3,5\}$.
    }
    \label{fig:method}
    \vspace{-0.5cm}
\end{figure}

Based on the idea, we present a new \proposed method to leverage \underline{R}ating \underline{O}rdinality in \underline{G}NN-based \underline{M}atrix \underline{C}ompletion (Figure \ref{fig:method}).
We convert the bipartite graph $G$ into a set of subgraphs $\{G_{t}\}_{t\in \mathcal{T}}$, where each subgraph $G_{t}$ includes edges with ratings greater than or equal to $t$.
Each subgraph's edges now represent whether the preference is stronger than a certain degree, providing multiple views generated based on inherent orders of rating types.
We apply GNN to each subgraph and make predictions by consolidating representations from the subgraphs.
Furthermore, we introduce a new interest regularization to assist preference learning from each subgraph.
As all ratings for an item are given after interactions driven by interest\footnote{In this paper, we use the term `interest' to denote what makes users try items and the term `preference' to denote users' explicit feedback given after trying items, i.e., users try interesting items and express their explicit preferences by ratings.}, there exists an underlying interest semantic regardless of the rating types. 
Based on this idea, we enable nodes in the subgraphs with different preference levels to share the same semantics of user interest.
We validate the validity of \proposed with extensive experiments on real-world datasets, and provide a detailed analysis showing the effectiveness of each proposed component.

\section{Methodology}
\subsection{Problem Formulation and Notations}
\label{sec:pre}
Given a partially observed user-item matrix $M$, where each entry $M_{ij}$ denotes the feedback from user $i$ on item $j$, matrix completion (MC) aims to predict the missing entries.
In practical applications, users typically rate only a small fraction of their purchases \cite{APPL14, KOTU2019343}, thus there inevitably exist interacted-but-unrated items.
Consequently, some observed entries have ratings from a set $\mathcal{R} = \{1,..., R\}$, while other entries are observed but their ratings remain unknown, which are denoted as $U$.
The matrix $M$ can be viewed as a user-item bipartite graph $G$, where nodes represent users and items, and edges represent observed user-item interactions \cite{inductiveMC_CIKM21, GCMC}.
Each edge connecting node $i$ and node $j$ has a weight of $M_{ij}$.

\subsection{\proposed}
\label{sec:method}

We present \proposed to exploit \underline{R}ating \underline{O}rdinality in \underline{G}NN-based \underline{MC}.
We focus on modeling the rating ordinality in GNN's message passing, unlike the prior methods treating rating types~independently.

\subsubsection{\textbf{Cumulative preference propagation}}
Employing the idea that converts an ordinal attribute into multiple binary attributes~\cite{frank2001simple}, we convert $G$ into a set of multiple unweighted subgraphs. 
Each subgraph contains all the nodes, but different edges according to the rating types.
Specifically, we generate a set of subgraphs $\{G_{t}\}_{t\in \mathcal{T}}$ by collecting edges with ratings greater than or equal to $t$ into $G_t$, where $\mathcal{T} \subseteq \mathcal{R}$ is the set of rating types used for the conversion.
Note that edges with unknown ratings (i.e., $U$) are not included~in~$G_{t}$.

Then, we apply GNN to each subgraph $G_t$ to exploit the graph structure by propagating representations on it.
Various GNNs can be employed here, and we follow the simple propagation rule of~\cite{he2020lightgcn, inductiveMC_CIKM21}:
\begin{equation}
\begin{aligned}
    e^{l+1}_t[i] = \sum_{j \in \mathcal{N}_t(i)} \frac{1}{c_{ij}} e^l_t[j],
\end{aligned}
\end{equation}
where $\mathcal{N}_t(i)$ denotes the set of node $i$'s neighbors in $G_t$, $c_{ij}=(|\mathcal{N}_{r}(i)| |\mathcal{N}_{r}(j)|)^{0.5}$ is the normalization constant, and $e^l_t[i]$ denotes node $i$’s representation at $l$-th layer.
It is worth noting that the propagation on each subgraph is independent, thus they can be efficiently computed in parallel. 
After $L$ propagation layers, we obtain the node representation from each subgraph as $h_t[i] = e^L_t[i]$.
Lastly, we make the final node representation $h[i]$ by aggregating the subgraph representations \cite{inductiveMC_CIKM21},~i.e.,~$h[i]=\sum_{t \in \mathcal{T}} h_t[i]$.

\textbf{Remarks.}
Unlike $G$ where edges correspond to individual rating types, edges in $G_t$ signify whether the ratings are higher than a certain degree.
This allows for considering multiple views generated based on inherent orders of rating types.
An interesting property of this approach is that edges with higher ratings are more frequently included in the subgraphs, which ensures their increased participation in the propagation process.
As a result, information associated with these edges is accumulated in $h[i]$ to a greater extent, emphasizing on users' stronger preferences based~on~the~rating~ordinality.

Alternatively, one can also consider emphasizing lower ratings by constructing $G_{t}$ with edges having ratings \textit{less than} or equal to $t$.
However, in our experiments, we consistently achieved better performance when focusing on higher ratings (reported in §3.2.2).

\subsubsection{\textbf{Interest regularization}}
Due to the limited number of ratings, $\{G_{t}\}_{t\in \mathcal{T}}$ has a high sparsity.
In particular, the subgraph with a stronger preference level has a higher sparsity, which hinders effective learning.
To cope with the sparsity problem, we introduce a new interest regularization that utilizes users' general interest information to guide the learning users'~preference~from~$\{G_{t}\}_{t\in \mathcal{T}}$.

Let $G_{I}$ denote the unweighted version of $G$, which includes all nodes and edges (i.e., both $\mathcal{R}$ and $U$).
In $G_{I}$, edges represent the existence of user-item interactions. 
By applying GNN, we can obtain valuable signals of what attracts the user's interest and makes them interact with items \cite{he2020lightgcn}.
Considering all ratings for an item are given after interactions driven by interest, there exists an underlying interest semantic regardless of the rating types.
We seek to leverage these underlying semantics to assist preference learning based on $\{G_{t}\}_{t\in \mathcal{T}}$. 
To this end, we make $\{h_{t}[i]\}_{t\in \mathcal{T}}$ to share the underlying semantic of interest $h_{I}[i]$ as follows:
\begin{equation}
\begin{aligned}
    \mathcal{L}_{IR} = \sum_{1 \leq i \leq N}\frac{1}{|\mathcal{T}|}\sum_{t\in \mathcal{T}} \big\lVert h_t[i] - h_I[i] \big\rVert_2
\end{aligned}
\end{equation}
where $h_{I}[i]$ denotes node $i$'s representation after applying $L$ propagation layers on $G_I$ (Eq.1), and $N$ is the total number of nodes. 

\subsubsection{\textbf{Bilinear decoder}}
Following \cite{GCMC, inductiveMC_CIKM21}, we use a bilinear decoder to predict ratings from the representations.
Given the final representation $h[i]$ of user $i$ and $h[j]$ of item $j$, the decoder produces a probability distribution over the rating types as follows:
\begin{equation}
\begin{aligned}
\hat{p}(M_{ij}=r) = \frac{\exp(z^r_{ij})}{\sum_{s \in \mathcal{R}}\exp(z^s_{ij})}, \quad z^r_{ij} = h[i]^{T} Q_{r} h[j],
\end{aligned}
\end{equation}
where $Q_{r}$ is a trainable parameter matrix. The final rating is computed by $\hat{M}_{ij} = \sum_{r \in \mathcal{R}} r \, \hat{p}(M_{ij}=r)$ \cite{GCMC}.

\subsubsection{\textbf{Model training}}
\proposed learns preference from ratings, with the regularization based on interest revealed from the existence of edges.
We use the rating prediction loss \cite{GCMC} along with the pair-wise ranking loss \cite{BPR}.
This pair-wise loss aids in capturing interest semantics by discriminating between observed and unobserved interactions, which can enhance the regularization effects.
Let $\mathcal{D}=\{(i,j,k) \mid M_{ij} \in \mathcal{R} \land M_{ik} \not\in \mathcal{R} \cup \{U\}\}$ denote the set of training instances, i.e., user $i$ gives rating to item $j$ but doesn't interact with item $k$.
The loss function is defined as follows:
\begin{equation}
\small
\begin{aligned}
\mathcal{L}=-\sum_{(i, j, k) \in \mathcal{D}} \left( 
\sum_{r\in \mathcal{R}} I(r=M_{ij}) \log p(\hat{M}_{ij}=r)
+
\alpha \log\sigma \left(o_{ij} - o_{ik} \right)
\right) 
\end{aligned}
\end{equation}
where the first term corresponds to the cross-entropy loss between the predictions and the ground-truth ratings \cite{GCMC, inductiveMC_CIKM21}, the second term is the pair-wise loss.
$I(\cdot)$ is the indicator function, $\alpha$ is a hyperparameter balancing the two terms, and $\sigma(\cdot)$ is the sigmoid function.
$o_{ij}$ denotes the interest score of user $i$ to item $j$.
We compute it by the sum of unnormalized scores, i.e., $o_{ij} = \sum_{r\in\mathcal{R}} z^r_{ij}$.
Finally, \proposed trains the model with the following loss function:
\begin{equation}
\begin{aligned}
\mathcal{L}_{ROGMC}= \mathcal{L} + \lambda \mathcal{L}_{IR}
\end{aligned}
\end{equation}
where $\lambda$ is a hyperparameter controlling the regularization effects.

\section{Experiments}
\subsection{Experiment Setup}
\noindent
\textbf{Datasets.}
We use three real-world datasets, ML-100K, ML-1M, and Eachmovie, widely used in recent work \cite{GCMC, inductiveMC_CIKM21, inductiveMC_ICLR20, christakopoulou2015collaborative, shi2020user}.
We use the 10-core setting \cite{he2020lightgcn, NGCF}, and data statistics are presented in Table \ref{tbl:statistic}.
\vspace{-0.3cm}
\begin{table}[h]
\centering
\renewcommand{\arraystretch}{0.50}
\renewcommand{\tabcolsep}{0.6mm}
  \caption{Data Statistics}
  \footnotesize
  \begin{tabular}{cccccc}
    \toprule
    Dataset & \#Users & \#Items & \#Ratings & Sparsity & Rating type \\
    \midrule
    ML-100K & 943 & 1,682 & 100,000 & 93.70\% & \{1,2,3,4,5\}\\
    ML-1M & 6,040 & 3,706 & 1,000,209 & 95.53\% & \{1,2,3,4,5\}\\
    Eachmovie & 72,916 & 1,628 & 2,811,983 & 97.63\% & \{1,2,3,4,5,6\}\\
    \bottomrule
  \end{tabular}
    \label{tbl:statistic}
    \vspace{-0.4cm}
\end{table}

\noindent
\textbf{Evaluation protocols.}
For each dataset, we randomly divide each user’s ratings into training/validation/test sets in an 80\%/10\%/10\% split.
We report the average root mean square error (RMSE) and its standard deviation from three independent runs.
To simulate the real-world scenarios having many interacted-but-unrated items \cite{KOTU2019343, APPL14}, we introduce additional experiment settings where ratings are given only a certain fraction of interactions (i.e., \textit{Rating-fracs}).
That is, in the 25\% setting, we keep the ratings for randomly selected 25\% of training interactions, and replace the ratings for the remaining interactions with unknown ratings (i.e., $U$).

\noindent
\textbf{Baselines.}
As the focus of this work is to develop a new approach to leverage rating ordinality for GNN, we focus on the comparison with the existing strategies for handling rating types, excluding other task-oriented designs (e.g., various input features for inductive prediction \cite{inductiveMC_CIKM21}).
Our baselines include a non-GNN method: \textbf{MF \cite{MF}}, and four GNN-based methods: \textbf{LGCN \cite{he2020lightgcn}}, \textbf{GCMC \cite{GCMC}}, \textbf{RGCN \cite{relation_GNN}}, and \textbf{EGCN \cite{edge_GNN}}.
LGCN treats all rating types equally.
Both GCMC and RGCN employ rating-specific transformations, but GCMC uses an additional dense layer. 
EGCN uses a unique edge embedding for each rating type \cite{edge_GNN}.
Note that the existing methods do not consider unknown ratings (i.e., $U$).
To ensure a fair comparison, we modify the GNN-based methods to treat $U$ as a separate relation type and report their performance as well.
The modified versions are denoted with a `+' symbol.

\noindent
\textbf{Experiment details.}
For all methods, we apply a grid search for hyperparameters and follow the search ranges reported in the existing work.
We also tried adjacent rating regularization \cite{inductiveMC_ICLR20}, but no significant improvement is~observed.
As this short paper focuses on showing the validity of our approach, we simply set $\mathcal{R}$ as $\mathcal{T}$.
We leave further exploration to find the optimal $\mathcal{T}$ for future study.

\subsection{Experiment Results}
\subsubsection{\textbf{Overall Evaluation}}
\label{subsec:exp_result}
Table \ref{tbl:main_result} presents the performance comparison on two different \textit{Rating-fracs}.
First, at \textit{Rating-frac}=100\%, GNN-based methods that treat rating types as independent relations (i.e., GCMC, RGCN, and EGCN) largely outperform LGCN that completely disregards rating types.
However, when \textit{Rating-frac}=25\%, the advantage of these methods diminishes, and they even fail to outperform LGCN on ML-100K.
This shows the effectiveness of the existing strategies for modeling rating types is limited, particularly when the training ratings are sparse.
Second, although the modified GNNs (denoted by `+') generally perform better than the original methods at \textit{Rating-frac}=25\%, these improvements are not always substantial. 
Lastly, \proposed consistently achieves the best performance among all competitors, with larger improvements when fewer ratings are available. 
These results show that directly modeling rating ordinality in GNN can indeed improve recommendation accuracy, and also support the validity of the proposed approach.

\begin{table}[t!]
\centering
\caption{RMSE results. $\textit{Gain}_{best}$ denotes the RMSE improvement of \proposed over the best~competitor.}
\renewcommand{\arraystretch}{0.9}
\renewcommand{\tabcolsep}{0.75mm}
\footnotesize
  \begin{minipage}[t]{1\linewidth}
  \centering
  \begin{tabular}{clcccccc}
    \toprule
    Rating-frac & Method & ML-100K & ML-1M & Eachmovie \\
    \midrule
    & MF & 0.9125 $\pm$ 0.0100 & 0.8994 $\pm$ 0.0018 & 1.2039 $\pm$ 0.0029 \\
    &LGCN+ & 0.9329 $\pm$ 0.0059 & 0.8983 $\pm$ 0.0048 & 1.2212 $\pm$ 0.0012 \\
    &GCMC & 0.8980 $\pm$ 0.0091 & 0.8446 $\pm$ 0.0063 & 1.1733 $\pm$ 0.0037 \\
    &GCMC+ & \underline{0.8935 $\pm$ 0.0142} & 0.8478 $\pm$ 0.0099 & 1.1601 $\pm$ 0.0102 \\
    100\%&RGCN & 0.9082 $\pm$ 0.0080 & \underline{0.8417 $\pm$ 0.0010} & \underline{1.1160 $\pm$ 0.0033} \\
    &RGCN+ & 0.9052 $\pm$ 0.0157 & 0.8613 $\pm$ 0.0161 & 1.1219 $\pm$ 0.0046 \\
    &EGCN & 0.8997 $\pm$ 0.0054 & 0.8540 $\pm$ 0.0029 & 1.1230 $\pm$ 0.0025 \\
    &EGCN+ & 0.8996 $\pm$ 0.0047 & 0.8531 $\pm$ 0.0033 & 1.1240 $\pm$ 0.0013 \\ 
    &\proposed & \textbf{0.8726 $\pm$ 0.0009} & \textbf{0.8340 $\pm$ 0.0035} & \textbf{1.1106 $\pm$ 0.0013} \\
    \cmidrule{2-5}
    &$\textit{Gain}_{best}$ & {0.0209} & {0.0077} & {0.0054} \\
    \cmidrule{1-5}
    & MF & 1.0870 $\pm$ 0.0087 & 0.9374 $\pm$ 0.0007 & 1.3287 $\pm$ 0.0058 \\
    &LGCN+ & \underline{0.9609 $\pm$ 0.0149} & 0.9222 $\pm$ 0.0012 & 1.2804 $\pm$ 0.0019 \\
    &GCMC & 0.9826 $\pm$ 0.0175 & 0.9158 $\pm$ 0.0151 & 1.2932 $\pm$ 0.0090 \\
    &GCMC+ & 0.9675 $\pm$ 0.0254 & \underline{0.8949 $\pm$ 0.0228} & 1.2611 $\pm$ 0.0062 \\
    25\%&RGCN & 0.9779 $\pm$ 0.0091 & 0.9135 $\pm$ 0.0025 & 1.2642 $\pm$ 0.0153 \\
    &RGCN+ & 0.9739 $\pm$ 0.0210 & 0.9122 $\pm$ 0.0063 & \underline{1.2145 $\pm$ 0.0015} \\
    &EGCN & 1.0365 $\pm$ 0.0378 & 0.9127 $\pm$ 0.0134 & 1.2400 $\pm$ 0.0022 \\
    &EGCN+ & 0.9744 $\pm$ 0.0146 & 0.9152 $\pm$ 0.0175 & 1.2250 $\pm$ 0.0019 \\
    &\proposed & \textbf{0.9317 $\pm$ 0.0025} & \textbf{0.8781 $\pm$ 0.0016} & \textbf{1.2039 $\pm$ 0.0013} \\
    \cmidrule{2-5}
    &$\textit{Gain}_{best}$ & 0.0292 & 0.0168 & 0.0106 \\
    \bottomrule
  \end{tabular}
  \end{minipage}
    \label{tbl:main_result} 
    \vspace{-0.4cm}
\end{table}

\subsubsection{\textbf{Ablation Study}}
\label{sec:ablation}
Table \ref{tbl:ablation} presents the performance with various ablations and design choices on ML-100K. 
First, in (a)-(c), we ablate CP and IR from \proposed.\footnote{Note that \proposed without CP and IR can be thought of as LGCN+. IR cannot be applied to LGCN+ alone, as it acts on multiple node representations.}  
We observe that both components are indeed effective, and the best performance is achieved by using both of them.
Notably, IR effectively improves preference learning from the sparse subgraphs, without leveraging any extra rating data.
Second, we compare other choices for constructing multi-view subgraphs.
(d) and (e) construct each subgraph $G_t$ by collecting edges with ratings exactly equal to $t$ and those with ratings less than or equal to $t$, respectively.\footnote{(d) can be seen as the integration of rating-wise propagation \cite{inductiveMC_CIKM21} within \proposed.}
We observe that directly leveraging ordinality shows better results than rating-wise propagation ((c) vs. (d)), and emphasizing higher ratings is more effective than focusing on lower ratings ((c) vs. (e)).
Lastly, (f) ablates the pair-wise loss, which is employed to enhance the effects of interest regularization.
This modification leads to a slight decline in the final performance.
These results collectively support the validity of each proposed component and our design choices.

\begin{table}[t]
\footnotesize
\centering
\renewcommand{\arraystretch}{0.5}
\renewcommand{\tabcolsep}{2.5mm}
  \caption{RMSE results for ablations. `CP' denotes cumulative preference propagation, `IR' denotes interest regularization.}
  \begin{tabular}{clcccc}
    \toprule
    \multicolumn{2}{l}{\textbf{Ablations}} & CP & IR & Rating-frac (100\% / 50\% / 25\%)\\
    \midrule
    (a)& LGCN+ & \xmark & \xmark & 0.9329 / 0.9448 / 0.9609\\
    (b)& & \checkmark & \xmark & 0.8967 / 0.9228 / 0.9467\\
    (c)& \proposed & \checkmark &\checkmark & 0.8726 / 0.9042 / 
    0.9317\\
    \midrule
    \midrule
    \multicolumn{4}{l}{\textbf{\proposed with other design choices}} & Rating-frac (100\% / 50\% / 25\%)\\
    \midrule
    (d)& \multicolumn{3}{l}{subgraph ($=$)} & 0.8821 / 0.9117 / 0.9472\\
    (e)& \multicolumn{3}{l}{subgraph ($\leq$)} & 0.8936 / 0.9165 / 0.9492\\
    (f)& \multicolumn{3}{l}{IR w/o pair-wise loss}  & 0.8835 / 0.9195 / 0.9438\\
    \bottomrule
  \end{tabular}
    \label{tbl:ablation}
    \vspace{-0.3cm}
\end{table}

\subsubsection{\textbf{Further Analysis}}
\label{sec:further}
\begin{figure}[t]
\centering
    \includegraphics[width=1.\linewidth]{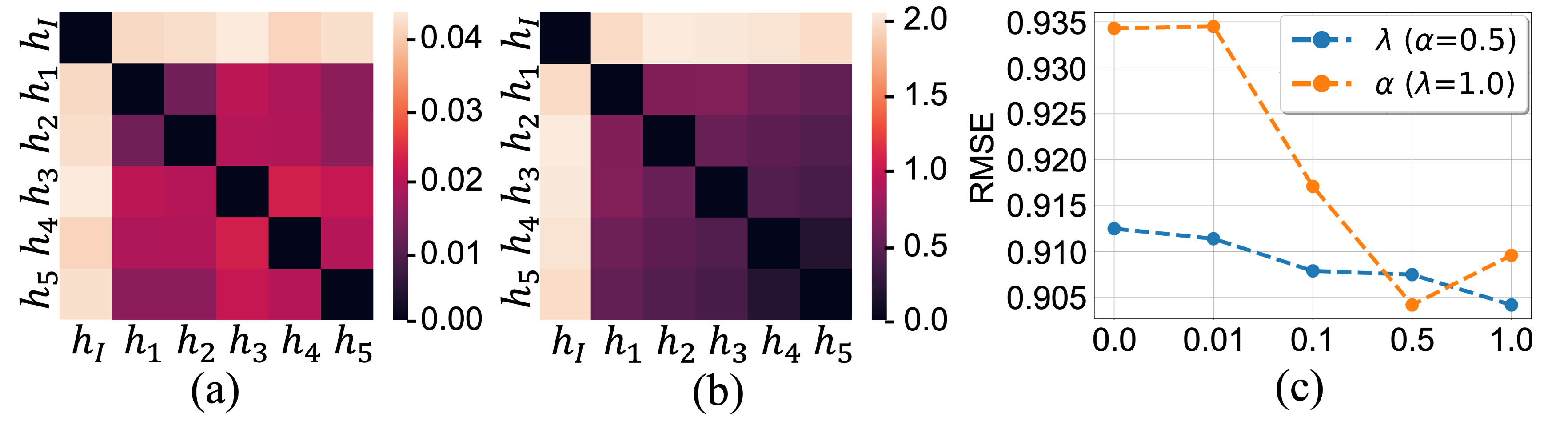}
    \caption{Effects of \proposed. (a-b) L2 distances between the averaged user representations across various rating types. (a) and (b) corresponds to the results of GCMC+ and \proposed, respectively. (c) RMSE with varying hyperparameters: $\lambda$ and $\alpha$. Results of \textit{Rating-frac}=50\% on ML-100K. }
    \label{fig:further_analysis}
    \vspace{-0.5cm}
\end{figure}

We provide further analysis to provide deeper insight for \proposed.
First, we investigate whether the rating ordinality is well captured by \proposed.
We compute the averaged user representation $AVG_{i}(h_t[i])$ for each rating type $t$, and compute the L2 distance between the averaged representations across the rating types.
Fig.\ref{fig:further_analysis}a presents the results of GCMC+ which shows the best performance on ML-100K, and Fig.\ref{fig:further_analysis}b presents the results of \proposed.
We observe that \proposed better encodes the inherent orders of ratings in its representation space compared to GCMC+.
For example, in terms of the degree of preference, rating 5 is closer to rating 4 than rating 1.
So, it is expected that the distance between $h_5$ and $h_4$ would be smaller than the distance between $h_5$ and $h_1$.
Such ordinal relationships are better reflected in \proposed, whereas GCMC+ often shows largely contradictory results to the ordinality.

Interestingly, in Fig.\ref{fig:further_analysis}b, representations related to stronger preferences (e.g., $h_4$ and $h_5$) generally show smaller distances with other representations.
We believe that it is a reasonable result considering that our cumulative preference propagation exploits edges with higher ratings more frequently so that stronger preferences are more emphasized.
Based on the observations, we conclude that \proposed better encodes the rating ordinality compared to the existing strategy that treats rating types as independent~relations.

Lastly, Fig.\ref{fig:further_analysis}c shows the effects of two hyperparameters:
$\lambda$ that controls the impacts of interest regularization, and $\alpha$ that balances the pair-wise loss with the rating prediction loss.
The best performance is achieved when $\lambda$ is around 1.0 and $\alpha$ is around 0.5.

\section{Conclusion}
\label{sec:conclusion}
We explore a new approach to exploit rating ordinality for GNN-based matrix completion.
Unlike the existing methods that treat rating types independently, \proposed directly utilizes the inclusion relationships according to the rating ordinality.
\proposed uses \textit{cumulative preference propagation} to consider multiple views generated based on the rating ordinality, along with \textit{interest regularization} to facilitate preference learning using the underlying interest semantics.
We validate the validity of \proposed with extensive experiments.

This short paper primarily explores the feasibility of leveraging rating ordinality for GNNs.
We expect that our findings can stimulate further research in this direction. 
In our future work, we will delve deeper into the applicability and scalability of our approach, particularly in the context of web-scale datasets.

\bibliographystyle{ACM-Reference-Format}
\bibliography{acmart}


\begin{thebibliography}{17}


\ifx \showCODEN    \undefined \def \showCODEN     #1{\unskip}     \fi
\ifx \showDOI      \undefined \def \showDOI       #1{#1}\fi
\ifx \showISBNx    \undefined \def \showISBNx     #1{\unskip}     \fi
\ifx \showISBNxiii \undefined \def \showISBNxiii  #1{\unskip}     \fi
\ifx \showISSN     \undefined \def \showISSN      #1{\unskip}     \fi
\ifx \showLCCN     \undefined \def \showLCCN      #1{\unskip}     \fi
\ifx \shownote     \undefined \def \shownote      #1{#1}          \fi
\ifx \showarticletitle \undefined \def \showarticletitle #1{#1}   \fi
\ifx \showURL      \undefined \def \showURL       {\relax}        \fi
\providecommand\bibfield[2]{#2}
\providecommand\bibinfo[2]{#2}
\providecommand\natexlab[1]{#1}
\providecommand\showeprint[2][]{arXiv:#2}

\bibitem[Berg et~al\mbox{.}(2017)]%
        {GCMC}
\bibfield{author}{\bibinfo{person}{Rianne van~den Berg}, \bibinfo{person}{Thomas~N Kipf}, {and} \bibinfo{person}{Max Welling}.} \bibinfo{year}{2017}\natexlab{}.
\newblock \showarticletitle{Graph convolutional matrix completion}.
\newblock \bibinfo{journal}{\emph{arXiv preprint arXiv:1706.02263}} (\bibinfo{year}{2017}).
\newblock


\bibitem[Chen and Peng(2018)]%
        {chen2018matrix}
\bibfield{author}{\bibinfo{person}{Shulong Chen} {and} \bibinfo{person}{Yuxing Peng}.} \bibinfo{year}{2018}\natexlab{}.
\newblock \showarticletitle{Matrix factorization for recommendation with explicit and implicit feedback}.
\newblock \bibinfo{journal}{\emph{Knowledge-Based Systems}}  \bibinfo{volume}{158} (\bibinfo{year}{2018}).
\newblock


\bibitem[Christakopoulou and Banerjee(2015)]%
        {christakopoulou2015collaborative}
\bibfield{author}{\bibinfo{person}{Konstantina Christakopoulou} {and} \bibinfo{person}{Arindam Banerjee}.} \bibinfo{year}{2015}\natexlab{}.
\newblock \showarticletitle{Collaborative ranking with a push at the top}. In \bibinfo{booktitle}{\emph{WWW}}. \bibinfo{pages}{205--215}.
\newblock


\bibitem[Frank and Hall(2001)]%
        {frank2001simple}
\bibfield{author}{\bibinfo{person}{Eibe Frank} {and} \bibinfo{person}{Mark Hall}.} \bibinfo{year}{2001}\natexlab{}.
\newblock \showarticletitle{A simple approach to ordinal classification}. In \bibinfo{booktitle}{\emph{12th European Conference on Machine Learning Freiburg}}. Springer.
\newblock


\bibitem[Guo et~al\mbox{.}(2014)]%
        {APPL14}
\bibfield{author}{\bibinfo{person}{Guibing Guo}, \bibinfo{person}{Jie Zhang}, \bibinfo{person}{Daniel Thalmann}, {and} \bibinfo{person}{Neil Yorke{-}Smith}.} \bibinfo{year}{2014}\natexlab{}.
\newblock \showarticletitle{Leveraging prior ratings for recommender systems in e-commerce}.
\newblock \bibinfo{journal}{\emph{Appl.}} (\bibinfo{year}{2014}).
\newblock


\bibitem[He et~al\mbox{.}(2020)]%
        {he2020lightgcn}
\bibfield{author}{\bibinfo{person}{Xiangnan He}, \bibinfo{person}{Kuan Deng}, \bibinfo{person}{Xiang Wang}, \bibinfo{person}{Yan Li}, \bibinfo{person}{Yongdong Zhang}, {and} \bibinfo{person}{Meng Wang}.} \bibinfo{year}{2020}\natexlab{}.
\newblock \showarticletitle{LightGCN: Simplifying and Powering Graph Convolution Network for Recommendation}. In \bibinfo{booktitle}{\emph{SIGIR}}.
\newblock


\bibitem[Koren et~al\mbox{.}(2009)]%
        {MF}
\bibfield{author}{\bibinfo{person}{Yehuda Koren}, \bibinfo{person}{Robert Bell}, {and} \bibinfo{person}{Chris Volinsky}.} \bibinfo{year}{2009}\natexlab{}.
\newblock \showarticletitle{Matrix factorization techniques for recommender systems}.
\newblock \bibinfo{journal}{\emph{Computer}} \bibinfo{volume}{42}, \bibinfo{number}{8} (\bibinfo{year}{2009}), \bibinfo{pages}{30--37}.
\newblock


\bibitem[Kotu and Deshpande(2019)]%
        {KOTU2019343}
\bibfield{author}{\bibinfo{person}{Vijay Kotu} {and} \bibinfo{person}{Bala Deshpande}.} \bibinfo{year}{2019}\natexlab{}.
\newblock \showarticletitle{Chapter 11 - Recommendation Engines}.
\newblock In \bibinfo{booktitle}{\emph{Data Science (Second Edition)} (\bibinfo{edition}{second edition} ed.)}. \bibinfo{publisher}{Morgan Kaufmann}, \bibinfo{pages}{343--394}.
\newblock
\showISBNx{978-0-12-814761-0}


\bibitem[Park et~al\mbox{.}(2023)]%
        {CPA-LGC}
\bibfield{author}{\bibinfo{person}{Jin-Duk Park}, \bibinfo{person}{Siqing Li}, \bibinfo{person}{Xin Cao}, {and} \bibinfo{person}{Won-Yong Shin}.} \bibinfo{year}{2023}\natexlab{}.
\newblock \showarticletitle{Criteria Tell You More than Ratings: Criteria Preference-Aware Light Graph Convolution for Effective Multi-Criteria Recommendation} \emph{(\bibinfo{series}{KDD '23})}. \bibinfo{pages}{1808–1819}.
\newblock
\showISBNx{9798400701030}


\bibitem[Rendle et~al\mbox{.}(2009)]%
        {BPR}
\bibfield{author}{\bibinfo{person}{Steffen Rendle}, \bibinfo{person}{Christoph Freudenthaler}, \bibinfo{person}{Zeno Gantner}, {and} \bibinfo{person}{Lars Schmidt-Thieme}.} \bibinfo{year}{2009}\natexlab{}.
\newblock \showarticletitle{BPR: Bayesian personalized ranking from implicit feedback}. In \bibinfo{booktitle}{\emph{UAI}}.
\newblock


\bibitem[Schlichtkrull et~al\mbox{.}(2018)]%
        {relation_GNN}
\bibfield{author}{\bibinfo{person}{Michael Schlichtkrull}, \bibinfo{person}{Thomas~N Kipf}, \bibinfo{person}{Peter Bloem}, \bibinfo{person}{Rianne van~den Berg}, \bibinfo{person}{Ivan Titov}, {and} \bibinfo{person}{Max Welling}.} \bibinfo{year}{2018}\natexlab{}.
\newblock \showarticletitle{Modeling relational data with graph convolutional networks}. In \bibinfo{booktitle}{\emph{European semantic web conference}}. Springer, \bibinfo{pages}{593--607}.
\newblock


\bibitem[Shen et~al\mbox{.}(2021)]%
        {inductiveMC_CIKM21}
\bibfield{author}{\bibinfo{person}{Wei Shen}, \bibinfo{person}{Chuheng Zhang}, \bibinfo{person}{Yun Tian}, \bibinfo{person}{Liang Zeng}, \bibinfo{person}{Xiaonan He}, \bibinfo{person}{Wanchun Dou}, {and} \bibinfo{person}{Xiaolong Xu}.} \bibinfo{year}{2021}\natexlab{}.
\newblock \showarticletitle{Inductive Matrix Completion Using Graph Autoencoder}. In \bibinfo{booktitle}{\emph{CIKM}}. \bibinfo{pages}{1609--1618}.
\newblock


\bibitem[Shi et~al\mbox{.}(2020)]%
        {shi2020user}
\bibfield{author}{\bibinfo{person}{Wenchuan Shi}, \bibinfo{person}{Liejun Wang}, {and} \bibinfo{person}{Jiwei Qin}.} \bibinfo{year}{2020}\natexlab{}.
\newblock \showarticletitle{User embedding for rating prediction in SVD++-based collaborative filtering}.
\newblock \bibinfo{journal}{\emph{Symmetry}} \bibinfo{volume}{12}, \bibinfo{number}{1} (\bibinfo{year}{2020}), \bibinfo{pages}{121}.
\newblock


\bibitem[Wang et~al\mbox{.}(2019)]%
        {NGCF}
\bibfield{author}{\bibinfo{person}{Xiang Wang}, \bibinfo{person}{Xiangnan He}, \bibinfo{person}{Meng Wang}, \bibinfo{person}{Fuli Feng}, {and} \bibinfo{person}{Tat-Seng Chua}.} \bibinfo{year}{2019}\natexlab{}.
\newblock \showarticletitle{Neural graph collaborative filtering}. In \bibinfo{booktitle}{\emph{SIGIR}}. \bibinfo{pages}{165--174}.
\newblock


\bibitem[Zhang et~al\mbox{.}(2022)]%
        {edge_GNN}
\bibfield{author}{\bibinfo{person}{Chengkun Zhang}, \bibinfo{person}{Hongxu Chen}, \bibinfo{person}{Sixiao Zhang}, \bibinfo{person}{Guandong Xu}, {and} \bibinfo{person}{Junbin Gao}.} \bibinfo{year}{2022}\natexlab{}.
\newblock \showarticletitle{Geometric Inductive Matrix Completion: A Hyperbolic Approach with Unified Message Passing}. In \bibinfo{booktitle}{\emph{WSDM}}.
\newblock


\bibitem[Zhang et~al\mbox{.}(2019)]%
        {zhang2019star}
\bibfield{author}{\bibinfo{person}{Jiani Zhang}, \bibinfo{person}{Xingjian Shi}, \bibinfo{person}{Shenglin Zhao}, {and} \bibinfo{person}{Irwin King}.} \bibinfo{year}{2019}\natexlab{}.
\newblock \showarticletitle{Star-gcn: Stacked and reconstructed graph convolutional networks for recommender systems}. In \bibinfo{booktitle}{\emph{IJCAI}}.
\newblock


\bibitem[Zhang and Chen(2020)]%
        {inductiveMC_ICLR20}
\bibfield{author}{\bibinfo{person}{Muhan Zhang} {and} \bibinfo{person}{Yixin Chen}.} \bibinfo{year}{2020}\natexlab{}.
\newblock \showarticletitle{Inductive matrix completion based on graph neural networks}. In \bibinfo{booktitle}{\emph{ICLR}}.
\newblock


\end{thebibliography}

s

\end{document}